\documentclass[10pt,twocolumn,letterpaper]{article}

\usepackage[pagenumbers]{cvpr}      

\usepackage[pagebackref,breaklinks,colorlinks,citecolor=cvprblue]{hyperref}
\usepackage{microtype}
\usepackage{graphicx}
\usepackage{booktabs} 
\usepackage{multirow}

\usepackage{amsmath}
\usepackage{amssymb}
\usepackage{mathtools}
\usepackage{amsthm}
\usepackage[table, dvipsnames]{xcolor}
\usepackage{colortbl}
\usepackage{caption}

\usepackage[capitalize,noabbrev]{cleveref}
\usepackage{tikz}

\theoremstyle{plain}

\theoremstyle{definition}

\theoremstyle{remark}

\definecolor{tabthird}{rgb}{1, 0.9, 0.9} 
\definecolor{tabsecond}{rgb}{1, 0.85, 0.7} 
\definecolor{beaublue}{rgb}{0.74, 0.83, 0.9} 
\definecolor{babyblue}{rgb}{0.54, 0.81, 0.94} 
\definecolor{lightblue}{HTML}{deeff5} 
\definecolor{lightlightblue}{rgb}{0.9, 0.9, 0.9} 

\newcommand{\bh}{\mathbf{h}}
\newcommand{\bq}{\mathbf{q}}
\newcommand{\bk}{\mathbf{k}}
\newcommand{\bv}{\mathbf{v}}

\newcommand{\T}{\top}
\newcommand{\xT}{\textlangle{}xT\textrangle{} }
\newlength\savewidth
\newcommand{\tablestyle}[2]{\setlength{\tabcolsep}{#1}\renewcommand{\arraystretch}{#2}\centering\footnotesize}

\usepackage[textsize=tiny]{todonotes}

\def\paperID{*****} 
\def\confName{CVPR}
\def\confYear{2024}

\title{\textit{x}T: Nested Tokenization for Larger Context in Large Images}

\author{Ritwik Gupta\textsuperscript{1}\thanks{Denotes co-first authorship. Co-first authors will prioritize their names on their resumes/websites.}~\qquad
Shufan Li\textsuperscript{2}\footnotemark[1]~\qquad
Tyler Zhu\textsuperscript{1,3}\footnotemark[1]~\\
Jitendra Malik\textsuperscript{1}\qquad
Trevor Darrell\textsuperscript{1\textdagger}\qquad
Karttikeya Mangalam\textsuperscript{1\textdagger}
\\ \\ \textsuperscript{1}Berkeley AI Research\qquad \textsuperscript{2}UCLA\qquad \textsuperscript{3}Princeton University \\
{\small correspondence to }{\tt\small ritwikgupta@berkeley.edu}
}

\begin{document}
\maketitle

\newcommand{\ourmethod}{\textit{xT}}
\newcommand\numberthis{\addtocounter{equation}{1}\tag{\theequation}}

\definecolor{cvprblue}{rgb}{0.21,0.49,0.74}

\begin{abstract}
Modern computer vision pipelines handle large images in one of two sub-optimal ways: down-sampling or cropping. 
These two methods incur significant losses in the amount of information and context present in an image.
There are many downstream applications in which global context matters as much as high frequency details, such as in real-world satellite imagery; in such cases researchers have to make the uncomfortable choice of which information to discard. 
We introduce \ourmethod, a simple framework for vision transformers which effectively aggregates global context with local details and can model large images end-to-end on contemporary GPUs. 
We select a set of benchmark datasets across classic vision tasks which accurately reflect a vision model's ability to understand truly large images and incorporate fine details over large scales and assess our method's improvement on them.
\ourmethod~is a streaming, two-stage architecture that adapts existing vision backbones and long sequence language models to effectively model large images without quadratic memory growth. We are able to increase accuracy by up to 8.6\% on challenging classification tasks and $F_1$ score by 11.6 on context-dependent segmentation on images as large as 29,000 x 29,000 pixels.

Code and pre-trained weights are available at \href{https://github.com/bair-climate-initiative/xT}{https://github.com/bair-climate-initiative/xT}.
\end{abstract}

\section{Introduction}
\label{sec:intro}

As camera technology has advanced, images have been getting increasingly larger over the past decade. Images captured by sensors on smartphones now capture images at 4K resolution (roughly 8.3M pixels) while professional DSLR cameras capture images at 8K resolution. Elsewhere, sensors on satellites and microscopes capture images with over a billion pixels.

Modern computer vision pipelines are limited by the memory in the systems they are trained upon, resulting in the creation of models that only operate on small images. Computer vision practitioners limit the size of images in two less-than-ideal ways: down-sampling or cropping. While these simple operations produce powerful models when measured against typical computer vision benchmarks, the loss of high frequency information or global context is limited for many real-world tasks.

Consider a video feed of a football game. Captured natively in 8K resolution, a model attempting to answer the question of where a player on the left side of the screen will pass the ball to on the right side of screen will not be able to reason over the entire image in one pass. The image, the downstream model, and all intermediate tensors cannot fit in the memory of modern, large VRAM GPUs. A common approach is to process the image by treating it as individual ``windows'', each fed through the model without sharing context, resulting in sub-optimal performance.

\begin{figure}[t]
    \centering
    \includegraphics[width=\columnwidth]{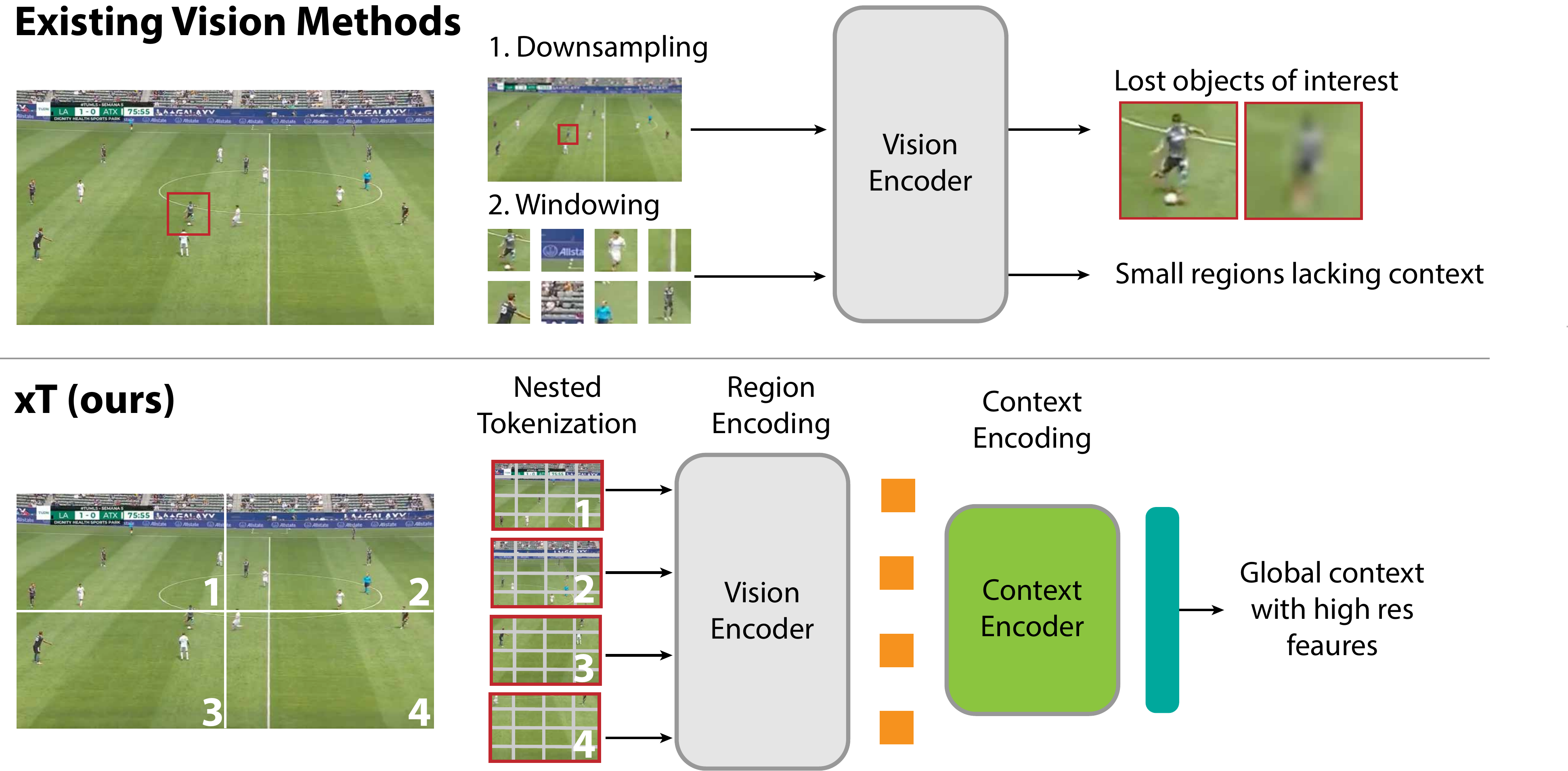}
    \caption{\ourmethod~allows large images to be modeled end-to-end on contemporary GPUs \textbf{without compromising on high frequency features or global context.}
    }
    \label{fig:teaser}
    \vspace{-1em}
\end{figure}

\begin{figure*}[t]
    \centering
    \includegraphics[width=\textwidth]{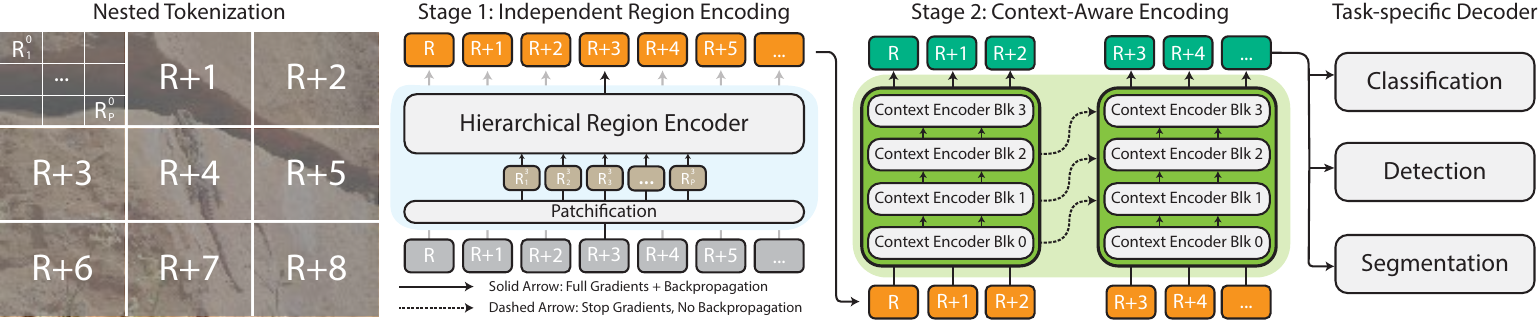}
    \caption{\textbf{Overview of our methodology.}
    \ourmethod~provides a way for existing vision backbones trained on small images to work effectively with large images. 
    The key is our \textbf{nested tokenization} of large images at multiple levels: at the region level as input $R^0, \dots, R^8$ ($R$, \dots, $R+8$ for readability) for the region encoders, and then at the patch level $R^i_0,\dots, R^i_{P-1}$ inside the encoders to understand local details. The image regions then undergo \textbf{independent, hierarchical encoding}, by passing through a vision backbone that serves as a region encoder. 
    Hierarchical region encoders result in down-sampled features which, when combined with \textbf{context encoders}, allows us to process more regions at once than typically possible.
    One such context encoder, Transformer-XL, is illustrated in Stage 2. It recurrently processes previous prior sequence tokens using cross attention, extending its context range significantly with depth.
    The resulting sequence has assimilated both local and global context and is finally fed to a \textbf{task-specific decoder.}}
    \label{fig:framework}
\end{figure*}

We introduce \ourmethod, a streaming, two-stage framework by which myopic vision backbones can effectively integrate local and global context over large images without incurring quadratic memory growth.
In particular, we tackle both issues of increasing GPU memory utilization and the integration of context across very large images. We achieve this by introducing token hierarchies to state-of-the-art vision backbones~\cite{liuSwinTransformerHierarchical2021, ryaliHieraHierarchicalVision2023} and imbuing the resulting local features with global context through the use of long-sequence models, such as Transformer-XL and Mamba~\cite{daiTransformerXLAttentiveLanguage2019, mamba}, obtained from the field of natural language processing. 
\ourmethod~matches and beats the performance of competitive large image architectures on multiple downstream tasks that require large visual contexts such as segmentation, detection, and classification. We demonstrate results on a variety of downstream tasks and achieve up to an 8.6\% gain in accuracy on classification tasks and an 11.6 increase in $F_1$ score on context-dependent segmentation.

\section{Related Works}
\label{sec:related}

\paragraph{Modeling large images}
Many prior works have attempted to model large images.
These approaches fall into one of two buckets: 1) multi-pass hierarchical or cascading approaches and 2) sliding windows combined with some suppression mechanism.
Dalal and Triggs famously utilized a sliding window approach for robust object recognition in their work on histograms of oriented gradients~\cite{dalalHistogramsOrientedGradients2005}.
R-CNN~\cite{girshickRichFeatureHierarchies2014} classified proposals generated by selective search in a cascade, resulting in a slow, albeit effective detector that worked for large images.
Gadermayr, et. al.~\cite{gadermayrCNNCascadesSegmenting2019} proposed a cascading convolutional neural network (CNN) where regions of interest (RoI) are first segmented using low resolution imagery.
These RoI are used to refine the segmentation mask using high resolution imagery.
~\cite{zhangMultiScaleVisionLongformer2021} implement a multi-scale vision transformer architecture which outputs a feature pyramid combined with a linear attention mechanism.

As demonstrated by prior work, there are inherent benefits to using hierarchical backbones when attempting to represent multi-scale features, so all of our experiments are done with them.
The \ourmethod~framework is agnostic to the style of vision backbone used for feature extraction. 

\paragraph{Assessing global understanding} 
Methods that aim to improve the handling of high-resolution or large images rarely benchmark their methods on real-world, large image datasets. For example, both \cite{zhangMultiScaleVisionLongformer2021} and \cite{yangGPViTHighResolution2022} are ViT architectures with non-global attention to better model large images. However, both works experiment on datasets such as ImageNet which is comprised of images of $224\times224$ pixels large.

Other evaluation methods rely on na{\"i}ve data processing techniques such as cropping medium-sized images from Cityscapes into smaller chunks and independently stitching the outputs together~\cite{guMultiScaleHighResolutionVision2021}, or center-cropping a $224\times224$ patch from a larger (yet objectively small) ImageNet image~\cite{wangPyramidVisionTransformer2021}. Ultimately, these approaches are sensitive to changes in test-time resolution, though representation learning methods exist to rectify this issue~\cite{reedScaleMAEScaleAwareMasked2023}.

Real-world classification and segmentation datasets that are heavily dependent on their surroundings, or in cases where the object that needs to be modeled is a small fraction of the overall image, would benefit from better integration of global context. Myopic vision encoders which utilize a windowing approach may fail at large objects in detection tasks as objects will span multiple windows.

We evaluate on real-world datasets such as xView3-SAR~\cite{paoloXView3SARDetectingDark2022}, a dataset where the average image is $29,400\times 24,400$ pixels large, and iNaturalist 2018~\cite{vanhornINaturalistSpeciesClassification2018}, where the entire $800\times 800$ image is utilized at once for evaluation. We also evaluate on the standard Cityscapes dataset for direct comparison on a common vision benchmark.

\paragraph{Learning global context}
Convolutional neural networks operate over images in sliding windows defined by the size of the kernels used for convolution.
However, convolutions do not inherently aggregate information from surrounding windows.
Following seminal work by Hubel and Wiesel~\cite{hubelReceptiveFieldsBinocular1962} and Koenderink and Van Doorn~\cite{koenderinkStructureLocallyOrderless1999} on feature pooling in the visual cortex, LeCun, et. al.~\cite{lecunHandwrittenDigitRecognition1989} introduced ``average pooling'' into CNNs as a context aggregation mechanism. Dilated convolutions as implemented by Yu and Kolun~\cite{yuMultiScaleContextAggregation2016} further increased the receptive field of CNNs.

Some models treat memory as an explicit, external component of the model. Neural Turing machines and memory networks ~\cite{gravesNeuralTuringMachines2014, westonMemoryNetworks2015} learn an access mechanism to read/write from a fixed-size memory matrix that is external to the backbone itself. More recently, MeMViT~\cite{wuMeMViTMemoryAugmentedMultiscale2022} extends this concept to sequences of images by caching and compressing activations from prior sequences with attention as a learned access mechanism.

Transformers and vision transformers~\cite{vaswaniAttentionAllYou2017, dosovitskiyImageWorth16x162020}, consisting of stacked attention layers~\cite{bahdanauNeuralMachineTranslation2016}, are able to maintain context over a fixed number of tokens. Prior works attempt to address limitations in attention via approaches such as factorization and adaptive masking~\cite{childGeneratingLongSequences2019, sukhbaatarAdaptiveAttentionSpan2019}. Notably, the Transformer-XL~\cite{daiTransformerXLAttentiveLanguage2019} efficiently passes context over a large number of tokens by recycling state over recurring input segments.

Recurrent neural networks (RNNs)~\cite{hopfieldNeuralNetworksPhysical1982, rumelhartLearningInternalRepresentations1987} addressed context over long sequences by recurrently passing hidden states over a fixed set of time steps but suffered from catastrophic foregetting~\cite{hochreiterUntersuchungenDynamischenNeuronalen1991}. Subsequent work on long short-term memory~\cite{hochreiterLongShortTermMemory1997} tackled this issue via gating. Structured state space models (SSMs)~\cite{s4} have re-emerged recently as a viable successor to RNNs. Of the SSM family of models, the Mamba~\cite{mamba} architecture presents a selection mechanisms for state spaces allowing it to generalize to large sequence lengths.

\ourmethod~is a streaming, two-stage architecture in which a powerful vision backbones extracts features from regions of a large image in batches. These features are streamed to a shallow context encoder which integrates global context across local features, effectively increasing the receptive field of the vision backbone across the entire image while staying within memory and parameter limits.

\section{Background}
\label{sec:background}
In this section, we briefly summarize the needed background for methods used in our work.

\subsection{Long-Context Models as Context Encoders}
\ourmethod~utilizes long-context models originally designed for text in order to mix information across large images.
These methods extend the context length beyond the typical limit of transformers. 
Below we briefly review two techniques which we build upon as our context encoders: Transformer-XL~\cite{daiTransformerXLAttentiveLanguage2019} and Mamba~\cite{mamba}.

\paragraph{Transformer-XL}
uses recurrence to pass prior information to future windows via prior hidden states.
This effect propagates through depth, so an $N$-layer transformer capable of taking a length $L$ sequence can be easily extended to handle a sequence of length $NL$.

Each hidden state $\bh_\tau^n$ of layer $n$ for sequence $\tau$ is computed from the previous layer hidden states $\bh_{\tau - 1}^{n-1}$ and $\bh_{\tau}^{n-1}$ as
\begin{align*}
    \tilde{\bh}_\tau^n &= [\text{SG}(\bh_{\tau - 1}^{n-1}) \circ \bh_\tau^{n-1}] \\
    \bq_\tau^n, \bk_\tau^n, \bv_\tau^n &= \bh_\tau^{n-1}W_q^\T, \tilde{\bh}_\tau^nW_k^\T, \tilde{\bh}_\tau^nW_v^\T \\
    \bh_\tau^n &= \text{Transformer}(\bq_\tau^n, \bk_\tau^n, \bv_\tau^n)  \numberthis \label{eq:txl}
\end{align*}
where $\text{SG}$ stands for a stop gradient. 
This is the same as the original Transformer, except that the keys and values $\bk_\tau^n, \bv_\tau^n$ are computed using the previous sequence's hidden state $\bh_{\tau-1}^{n-1}$ in addition to the current sequence's hidden state $\bh_\tau^{n-1}$ using cross attention. 
This mechanism allows for the recurrence of the hidden states $\bh_\tau^n$ across layers.
The application of a stop gradient between sequences lets information be propagated without suffering the memory costs incurred with full sequence backpropagation.

\paragraph{State Space Models}
State space models \cite{s4,s4nd} have been re-discovered recently as a potential replacement for transformers in long-sequence modeling. These models can be formulated as ordinary differential equations of the form
\begin{align*}
    \frac{dh(t)}{dt} &= Ah(t) + Bx(t) \\
    y(t) &= Ch(t) + Dx(t) \numberthis \label{eq:ssm}
\end{align*}
where $x(t)\in \mathbb{R}$ is the input signal and $y(t)\in \mathbb{R}$ is the output signal. Practically, this is computed through a discretization of the ODE via the zero-order hold (ZOH) rule: 
\begin{align*}
\bar{A} &= \exp(\Delta A) \\
\bar{B} &= (\Delta A)^{-1} (\exp(\Delta A) - I) \cdot \Delta B \\
h(t) &= \bar{A}h_{t-1} + \bar{B}x_t \numberthis \label{eq:zoh_ht}\\
y_t &= Ch_t \numberthis \label{eq:zoh_yt}
\end{align*}
Mamba \cite{mamba} is a new state space model that introduces a selective scan mechanism that allows time-varying parameterizations. Mamba theoretically carries context across very long sequences without any loss in accuracy and is implemented efficiently using custom CUDA kernels.

\subsection{Linear attention mechanism}
A standard transformer block with multi-headed self attention requires quadratic memory with respect to sequence length $L$ for fully global context. 
This is not ideal in the face of limited GPU memory. 
HyperAttention~\cite{han2023hyperattention}  is an attention mechanism with near-linear complexity with respect to sequence length.
It reduces the complexity of naive attention by first finding large entries of the attention matrix using Locality Sensitive Hashing (LSH). 
These dominant entries, combined with another randomly sampled subset from the matrix, are then used to approximate output of naive attention. 
This approach is particular helpful when the long range context correspondences are sparse.

\section{Methodology}
\label{sec:methodology}

Our goal for \ourmethod~is to demonstrate a simple framework for allowing existing methods to process large images in a memory efficient and context-preserving manner.
We achieve this through a streaming, two-stage architecture.
First, images are tokenized in a multi-stage hierarchy (Section~\ref{subsec: nest-token}) into regions. These regions are encoded by a powerful, but myopic, vision encoder in batches (at worst case, in serial) (Section~\ref{subsec: region}), resulting in a set of local features for the entire image.
A lightweight context encoder then integrates information across all local features, culminating in a feature vector which contains contextualized global semantics over the entire image (Section~\ref{subsec: context}). This feature vector is then used for task-specific decoding.
Our overall pipeline is illustrated in Figure~\ref{fig:framework}.

\subsection{Nested Tokenization}
\label{subsec: nest-token}

Given a large input image of shape $\alpha H\times \beta W$, we first subdivide the image into $H\times W$ regions so that our region encoder can adequately process them. 
Each region $R^i$ is further patchified into $P$ patches, $R^i_0, \dots, R^i_{P-1}$, by the region encoder backbone in order to extract features for each region. 
The regions are non-overlapping and zero-padded in instances when the region size, $H \times W$, does not evenly divide the image size.

Typically our images and regions are square, so we use a simplified notation to denote our pipeline parameters.
We refer to a pipeline which receives images of size $\alpha R\times \alpha R$ and subdivides them into $R\times R$ regions as an $\alpha R/R$ setup.
Standard setups are $512/256$, or $4096/512$, in which we split our image into $2\times 2$ and $8\times 8$ tiles respectively.

\subsection{Region Encoder}
\label{subsec: region}

The region encoder is any vision model which has been previously trained on small images small images $H\times W$, usually $224\times 224$ or $256\times 256$.
The region encoder independently generates feature maps for each region $R^i_{1...P}$.
In our experiments, we utilize vision transformers which output a shorter sequence length than which is input to them. These sequence lengths are less than the equivalent length produced by isotropic ViTs~\cite{dosovitskiyImageWorth16x162020}. 
In this setup, we are able to effectively handle images with an increased number of regions, as our sequence length is reduced by $4\times$ or greater.
However, we also demonstrate the effectiveness of \ourmethod~with a CNN-based encoder in the Appendix.

We stream regions through the region encoder in batches when GPU memory allows. However, either when the image is too large such that all of its constituent regions cannot fit into GPU memory, or when the regions themselves are too large, we process the image sequentially.
The features generated from this process contain information limited to each region and are concatenated in row-major order to form a sequence.


\subsection{Context Encoder}
\label{subsec: context}

The context encoder is a lightweight sequence-to-sequence model that is able to effectively attend to long sequences. In \ourmethod, the context encoder plays the key role of disaggregating global context across local features extracted using the region encoder.
Critically, we constrain the context encoder to have a near-linear memory and time cost. 
This design allows \ourmethod~to ``see'' many other regions of the large image which otherwise would not be feasible with the naive usage of a vision model.
\ourmethod's usage of a context encoder significantly extends the receptive field of existing vision models with a marginal increase in memory cost and the number of parameters.

Our method acts primarily in two settings: when our sequence of region features fits entirely within the context encoder's context length, and when it does not. 
In the first, we simply process everything at once.
We use standard 2D positional embeddings which are added to the nested region features.

We experiment with three context encoders with linear ``attention'' mechanisms: a LLaMA-style~\cite{touvronLLaMAOpenEfficient2023} architecture using HyperAttention (referred to as \textit{Hyper}), and Mamba. These two settings are called \emph{\xT Hyper} and \emph{\xT Mamba}, where \xT is an operator joining the choice of region encoder with context encoder.

When our input sequence does not fit in the entire context length, we need to additionally  compress our regions to maintain contextual information for future regions.
We experiment with a derivative of Transformer-XL that utilizes HyperAttention, a form of linear attention, and absolute positional embeddings instead.
We denote this setting as \emph{\xT XL}.
At the time of writing, Mamba does not have an efficient kernel for implementing an XL-style recurrence, so we cannot apply Mamba to this setting.

\paragraph{Effective Receptive Field Calculations}
Vision models used as-is are able to have an effective receptive field across the entire image but have zero reception on large images as they simply run out of memory. We aim to demonstrate the effective receptive field of \ourmethod's two-stage architecture. In summary, the receptive field of \ourmethod~is global but with minor aliasing artifacts due to an effective convolution over the image with the region encoder.

We provide calculations of the effective receptive field for the Swin \textlangle{}xT\textrangle{} XL setup. 
This is further ablated in \Cref{sec:ablation-erf}.

In the context encoder, we concatenate the features of $C$ regions into one ``chunk'' as our features are both smaller than the inputs from the hierarchical regional encoder. 
The Transformer-XL context encoder additionally has reduced attention memory requirements from using HyperAttention, further improving the number of regions that fit into one ``chunk''.
Each region attends to all other regions in this chunk, and they also have access to the previous chunk's hidden states for more context flow through the model. 
Context scales as a function of depth in Transformer-XL.

Consequently, we can calculate the context enhancement we achieve beyond just a standard model query on a small image. 
If we use a region encoder with input size $R$ (typically 256) and receive a large image of size $\alpha R\times \beta R$, then we will have a total of $\alpha\beta$ regions available for our context encoder.
Then we increase our context from increased chunk sizes by a factor of $C$, and also increase it from our recurrent memory by a factor of $N$, the depth of the context model. These values are calculated in \Cref{tab:calculation_erf} and further visualized in \Cref{fig:erf}.

In total, our context is multiplied by $\alpha\beta NC$.
However, note that there is a trade-off between $\alpha\beta$ and $C$, as increasing the size of our input image limits the chunk sizes that we can create from the region features. 

\setlength{\arrayrulewidth}{0.8pt}
\begin{table}
    \centering
    \resizebox{\columnwidth}{!}{%
    \begin{tabular}{l|c|c|c|r}
    Model & Input (px) & Region (px) & XL Layers & Context \\ \hline
    Swin-B & 256 & 256 & - & 65,536 \\
    Swin-B & 512 & 512 & - & 65,536 \\
    \rowcolor{lightblue}    
    Swin-B \xT XL & 512 & 256 & 1 & 131,072 \\
    \rowcolor{lightblue}
    Swin-B \xT XL & 512 & 256 & 2 & 196,608 \\
    \rowcolor{lightblue}
    Swin-B \xT XL & 4096 & 256 & 2 & 786,432
    \end{tabular}
    }
    \caption{The effective context length (in pixels) is calculated for Swin-B versus Swin-B \xT XL. Context is fixed for Swin while it grows as a function of region size and the number of Transformer-XL layers for Swin \xT XL.}
    \label{tab:calculation_erf}
\end{table}

\section{Experiments}
\label{sec:experiments}

\paragraph{Settings}
We aim to demonstrate the efficacy of \ourmethod~across the most common vision tasks: classification, detection, and segmentation. To do so, we pick a representative and challenging benchmark dataset in each domain for experimentation.

We focus on iNaturalist 2018~\cite{vanhornINaturalistSpeciesClassification2018} for classification, xView3-SAR~\cite{paoloXView3SARDetectingDark2022} for segmentation, and Cityscapes~\cite{cordtsCityscapesDatasetSemantic2016} for detection. Since iNaturalist 2018 is a massive dataset, we focus on the \textit{Reptilia} super-class, the most challenging subset available in the benchmark~\cite{vanhornINaturalistSpeciesClassification2018}. xView3-SAR is a difficult segmentation dataset for two reasons: it is comprised of extremely large images that are 29,400$\times$24,400 pixels large on average, and the objects in the dataset are heavily influenced by their non-local surroundings. Lastly, Cityscapes is used for evaluation of detection and to serve as a common baseline across prior work.

\paragraph{Metrics}
We measure top-1 accuracy for classification. For detection, we measure mean average precision (mAP) along with mAP\textsubscript{Large}. mAP\textsubscript{Large} is of specific interest since large objects in large images cross multiple region boundaries, resulting in a challenging training and evaluation problem. Segmentation is measured in two ways: aggregate $F_1$ score for segmentation of any object in the image, and close-to-shore $F_1$ score for objects that are ``close to the shoreline". xView3-SAR is comprised of synthetic aperture radar images which demonstrate unique artifacts around busy areas not found in typical vision benchmarks. Particularly, objects within 2km of the shoreline are impacted heavily by a shoreline that may not be visible in crops around the object.

\subsection{Classification}
\label{sec:cls}

\setlength{\arrayrulewidth}{0.8pt}
\begin{table}
    \centering
    \resizebox{\columnwidth}{!}{%
    \begin{tabular}{l|c|r|rr}
    Model & Top-1 Acc & Size(s) & Param & Mem (GB)\\ 
    \specialrule{.15em}{.075em}{.075em} 
    Swin-T    & 53.76 & 256       & 31M & 0.30 \\
    \rowcolor{lightblue}
    Swin-T \textlangle{}xT\textrangle{} Hyper & 52.93 & 256/256 & 47M & 0.31 \\
    \rowcolor{lightblue}
    Swin-T \textlangle{}xT\textrangle{} Hyper & 60.56 & 512/256 & 47M & 0.29 \\
    \rowcolor{lightblue}
    Swin-T \textlangle{}xT\textrangle{} XL & 58.92 & 512/256 & 47M & 0.17  \\
    \rowcolor{lightblue}
    Swin-T \textlangle{}xT\textrangle{} Mamba & \textbf{61.97} & 512/256 & 44M & 0.29 \\
    \hline
    Swin-S    & 58.45 & 256       & 52M & 0.46 \\
    \rowcolor{lightblue}
    Swin-S \textlangle{}xT\textrangle{} Hyper & 57.04 & 256/256 & 69M & 0.46\\
    \rowcolor{lightblue}
    Swin-S \textlangle{}xT\textrangle{} Hyper & \textbf{63.62} & 512/256 & 69M & 0.46\\
    \rowcolor{lightblue}
    Swin-S \textlangle{}xT\textrangle{} XL & 62.68 & 512/256 & 69M & 0.23 \\
    \rowcolor{lightblue}
    Swin-S \textlangle{}xT\textrangle{} Mamba\textbf{*} & - & - & - & -\\
    \hline
    Hiera-B    & 48.60 & 224       & 54M & 0.26 \\
    Hiera-B+   & 50.47 & 224       & 73M & 0.33 \\
    \rowcolor{lightblue}
    Hiera-B \textlangle{}xT\textrangle{} Hyper & \textbf{57.20} & 448/224 & 70M & 0.21 \\
    \hline
    Swin-B    & 58.57 & 256       & 92M & 0.50 \\
    \rowcolor{lightblue}
    Swin-B \textlangle{}xT\textrangle{} Hyper & 55.52 & 256/256 & 107M & 0.61 \\
    \rowcolor{lightblue}
    Swin-B \textlangle{}xT\textrangle{} Hyper & \textbf{64.08} & 512/256 & 107M & 0.74\\
    \rowcolor{lightblue}
    Swin-B \textlangle{}xT\textrangle{} XL & 62.09 & 512/256 & 107M & 0.39 \\
    \rowcolor{lightblue}
    Swin-B \textlangle{}xT\textrangle{} Mamba & 63.73 & 512/256 & 103M & 0.58 \\
    \hline
    Swin-L    & 68.78 & 256       & 206M & 0.84 \\
    \rowcolor{lightblue}
    Swin-L \textlangle{}xT\textrangle{} Hyper & 67.84 & 256/256 & 215M & 1.06 \\
    \rowcolor{lightblue}
    Swin-L \textlangle{}xT\textrangle{} Hyper & 72.42 & 512/256 & 215M & 1.03 \\
    \rowcolor{lightblue}
    Swin-L \textlangle{}xT\textrangle{} XL & \textbf{73.47} & 512/256 & 215M & 0.53 \\
    \rowcolor{lightblue}
    Swin-L \textlangle{}xT\textrangle{} Mamba & 73.36 & 512/256 & 212M & 1.03 \\
    
    \end{tabular}
    }
    \caption{\textbf{Comparison to prior methods on iNaturalist Reptilia Classification.} 
    \colorbox{lightblue}{Our methods}~improve on prior works significantly, showing that previous methods still fail to integrate global context.
    Memory is per base region size, which is fixed for each comparison.
    \textbf{*}Custom Mamba CUDA kernels are incompatible with Swin-S at this time.}
    \label{tab:results_inat}
\end{table}

We utilize the SwinV2~\cite{liu2021swinv2} and Hiera~\cite{ryaliHieraHierarchicalVision2023} families of hierarchical vision models as the region encoders for the classification experiments. All variants of Swin---tiny, small, base, and large---are pretrained on the ImageNet-1k~\cite{russakovsky2015imagenet} dataset. Both Hiera-B and Hiera-B+ are initialized with MAE~\cite{heMaskedAutoencodersAre2022} pre-trained/ImageNet-1k fine-tuned weights. Two layers of \textit{Hyper} or four layers of Mamba are used as the context encoder, intialized randomly.

We train end-to-end on the Reptilia subset of iNaturalist 2018 for 100 epochs using the AdamW optimizer ($\beta_1=0.9$, $\beta_2=0.999$, $\epsilon=1\times10^{-8}$) using cosine learning rate decay schedule. Swin-T, Swin-S, Hiera-B/+, and their \ourmethod~variants use a base learning rate of $1\times10^{-4}$ while Swin-B, Swin-L, and their \ourmethod~variants use a base learning rate of $1\times10^{-5}$. \ourmethod's nested tokenization scheme is represented as two values, e.g. 512/256, where the first value is the size of the regions extracted from the input image and the second value is the input size expected by the region encoder.

\Cref{tab:results_inat} contains results for variants of Swin, Hiera, and their \ourmethod~variants on iNaturalist-Reptilia. Particularly, we demonstrate \ourmethod's results as a function of region/input size and type of context encoder. \ourmethod~outperforms their comparable baselines by up to 8.6\% in top-1 accuracy.

\begin{figure}
    \centering
    \includegraphics[width=\columnwidth]{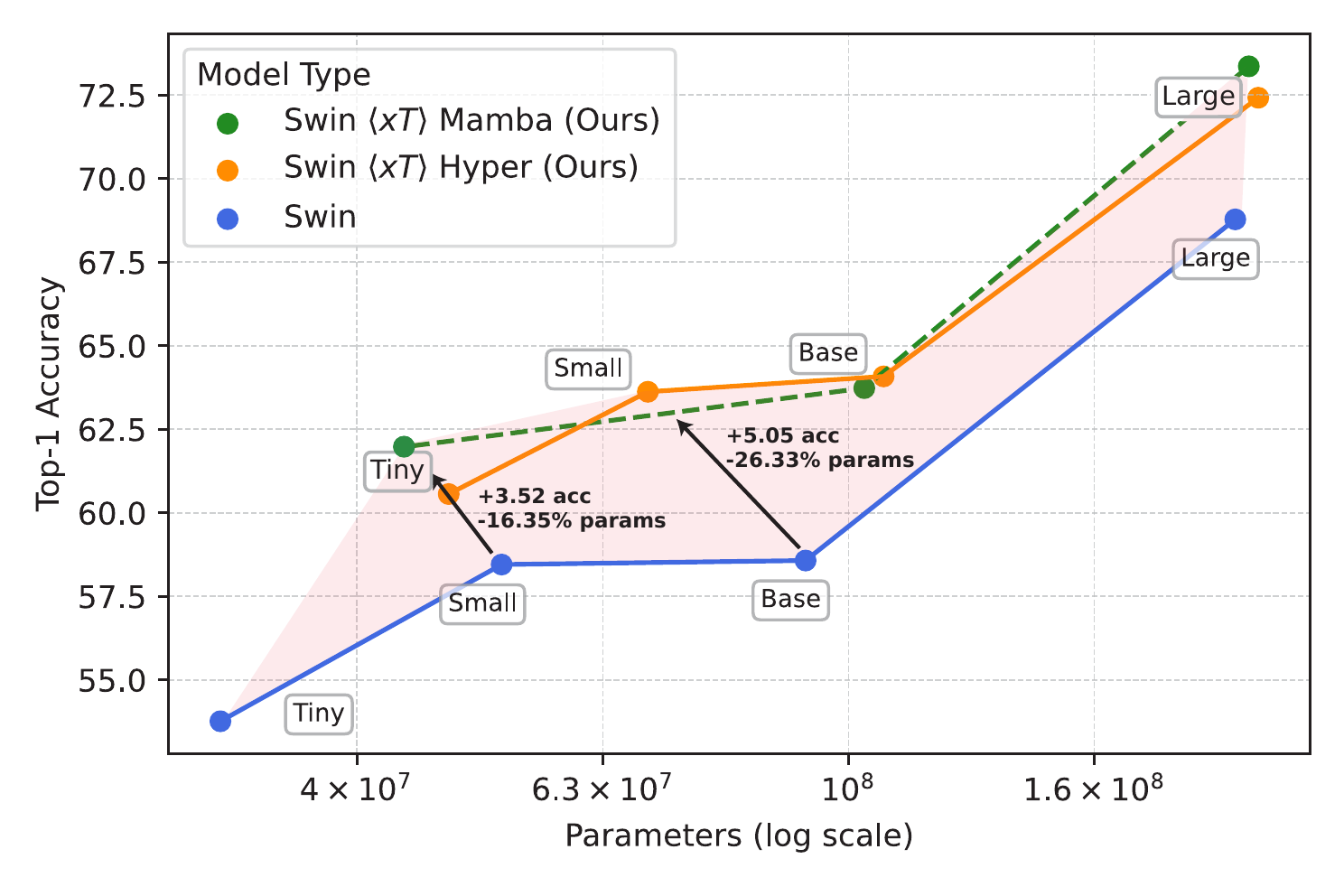}
    \caption{\textbf{\ourmethod~improves upon the accuracy-parameter frontier of existing methods for high-resolution classification on iNaturalist} using nested tokenization on top of the origin architecture for larger context. 
    See Table~\ref{tab:results_inat} for detailed comparisons.
    }
    \label{fig:frontier}
    \vspace{-1em}
\end{figure}

We show in \Cref{fig:frontier} that \ourmethod~sets a new accuracy-parameter frontier when compared against existing methods.

\subsection{Segmentation on xView3-SAR}

We evaluate \ourmethod~on the xView3-SAR~\cite{paoloXView3SARDetectingDark2022} dataset, a large, real-world dataset of satellite imagery for the task of detecting dark vessels.
The average image size is $29,400 \times 24,400$ pixels large, so accurate detections require using long-range geospatial context to make decisions.
We calculate two metrics on xView3-SAR: the aggregate and overall detection $F_1$ scores which reflect general task proficiency, and of most importance, the close-to-shore detection $F_1$, which requires detecting vessels close to the shore by using predominant shoreline information.

We adopt the same setup as prior methods~\cite{seferbekovXView32ndPlace2022}, tackling the problem as a segmentation and regression problem using a standard encoder-decoder architecture for dense prediction.
In our case, we adopt Swin Transformer v2~\cite{liu2021swinv2} pretrained on ImageNet-1k as our region encoder and use a UNet-style decoder~\cite{ronneberger2015u}.
We regress the centers of objects as Gaussians and vessels as binary masks.

We test Swin-T, Swin-S, and Swin-B, and their \ourmethod-variants using Transformer-XL as a context encoder with $N=4$ layers.
The models are trained over encoder input sizes of $512$ and $1024$, which \ourmethod~effectively boosts to 4096. 
At inference time, we take overlapping crops of the same size across the image and combine our predictions with post-processing.
We sweep our models over learning rates of $\{3\times10^{-5}, 10^{-4}, 3\times10^{-4}, 10^{-3}, 3\times10^{-3}\}$ using AdamW (with the same hyperparameters as in \Cref{sec:cls}) and report the validation numbers in Table~\ref{tab:results_xview}.

While prior works tackle xView3 using smaller models like CNNs, ours opts to adapt transformers for the task, which is an under-unexplored space for the type of images present. 
Therefore, larger models are not always better, as observed in Table~\ref{tab:results_xview}. 
However, \ourmethod~always outperforms the corresponding non-context model, beating baselines by up to 13.4 points on the overall $F_1$ detection score and 6.0 points on the close-to-shore $F_1$ score, signaling \ourmethod's ability to improve models on large context tasks. 
Due to our streaming, two-stage architecture with nested tokenization, the memory utilized per region is much lower as we see multiple regions at once per pass while only introducing $5-10\%$ more parameters, .

\setlength{\arrayrulewidth}{0.8pt}
\begin{table}
    \centering
    \resizebox{\columnwidth}{!}{%
    \begin{tabular}{l|c@{\hspace{0.7em}}c@{\hspace{0.7em}}c|rr|r}
    
    Model & Shore$\uparrow$ & Agg$\uparrow$ & $F_1$ Score$\uparrow$ & Param & Mem$\downarrow$ & Input Size(s) \\ \hline

    Swin-T & 50.0 & 47.6 & 67.8 & 32.4 & 1.24 & 512 \\
    Swin-T & 51.6 & 53.2 & 76.8 & 32.4 & 5.30 & 1024 \\
    \rowcolor{lightblue}
    Swin-T \xT XL  & 47.5 & 49.4 & \textbf{81.2} & 36.9 & \textbf{0.47} & 4096/512 \\
    \rowcolor{lightblue}
    Swin-T \xT XL& \textbf{56.0} & \textbf{54.8} & 78.2 & 36.9 & 1.65 & 4096/1024 \\
    \hline
    Swin-S & 46.1 & 44.9 & 67.7 & 53.7 & 1.84 & 512 \\
    Swin-S & 41.2 & 43.8 & 71.0 & 53.7 & 7.24 & 1024 \\
    \rowcolor{lightblue}
    Swin-S \xT XL & 50.2 & 48.1 & 75.3 & 58.3 & \textbf{0.54} & 4096/512  \\
    \rowcolor{lightblue}
    Swin-S \xT XL & \textbf{52.8} & \textbf{55.1} & \textbf{78.8} & 58.3 & 2.24 & 4096/1024 \\
    \hline
    Swin-B & 50.2 & 51.6 & 72.1 & 92.7 & 2.36 & 512 \\
    Swin-B & \textbf{54.4} & \textbf{54.7} & 75.8 & 92.7 & 9.65 & 1024 \\
    \rowcolor{lightblue}
    Swin-B \xT XL & 52.4 & 51.5 & 76.4 & 97.4 & \textbf{0.70} & 4096/512\\
    \rowcolor{lightblue}
    Swin-B \xT XL & 51.0 & 50.8 & \textbf{77.2} & 97.4 & 2.82 & 4096/1024
    \end{tabular}%
    }
    \caption{\textbf{Comparison to prior methods on xView3-SAR detection.} 
    We evaluate our methods on xView3-SAR, a dataset for dark vessel detection from satellite imagery.
    Our method improves on prior state of the art hierarchical transformers significantly while introducing few extra parameters and using less memory per region due to our efficient context encoder.
    This shows that prior works, expectedly, are unable to model long range visual contexts.}
    \label{tab:results_xview}
\end{table}

\begin{table}
    \centering
    \resizebox{\columnwidth}{!}{%
    \begin{tabular}{l|c|c}
    Model & mAP & Input Size(s) \\
    \specialrule{.15em}{.075em}{.075em}
    
    Swin-B-DetINO     & OOM & 2048 \\
    
    \rowcolor{lightblue}
    Swin-B-DetINO \xT XL  & \textbf{43.0} & 2048/512 \\
    \end{tabular}
    }
    \caption{\textbf{Comparison with Swin-B on the Cityscapes object detection task.} Swin-B is unable to model Cityscapes images in their entirety within the memory of an 80GB A100.}
    \vspace{-1em} 
    \label{tab:results_cityscapes}
\end{table}

\begin{table*}[!t]
\centering
\subfloat[
\textbf{Context Encoder Backbone.}
ViT and Hyper perform similarly without cross chunk context, while Mamba performs most efficiently.
\label{tab:ablation_context_backbone}
]{
\centering
\begin{minipage}{.25\linewidth}
    \centering
    \tablestyle{4pt}{1.05}
    \begin{tabular}{lcrr}
        context & acc & param & mem \\
        \specialrule{.15em}{.075em}{.075em} 
        ViT & 63.26 & 107M &0.73\\
        \rowcolor{lightblue}
        Hyper & \textbf{64.08}  & 107M & 0.74\\
        \rowcolor{lightblue}
        Mamba & 63.73 & 103M & 0.58%
    \vspace{3em}
    \end{tabular}
\end{minipage}%
}
\hspace{1.2em}
\subfloat[
\textbf{Context Encoder Depth.} Larger region encoders benefit from deeper context encoders.
We choose $N=2$ for our default depth.
\label{tab:context_encoder_depth}
]{
\centering
\begin{minipage}{0.33\linewidth}{\begin{center}
\tablestyle{4pt}{1.05}
\begin{tabular}{ccc}
model & depth & acc \\
\specialrule{.15em}{.075em}{.075em} 
\multirow{2}{*}{Swin-T} & 1 & \textbf{86.38} \\ 
                        & \cellcolor{lightblue}2 & \cellcolor{lightblue}85.09 \\ \hline
\multirow{2}{*}{Swin-S} & 1 & \textbf{88.26} \\ 
                        & \cellcolor{lightblue}2 & \cellcolor{lightblue}88.03 \\ \hline                            
\multirow{2}{*}{Swin-B} & 1 & 89.08 \\ 
                        & \cellcolor{lightblue}2 & \cellcolor{lightblue}\textbf{90.73} \\ \hline
\multirow{2}{*}{Swin-L} & 1 & 91.67 \\ 
                        & \cellcolor{lightblue}2 & \cellcolor{lightblue}\textbf{94.48}
\end{tabular}%
\end{center}}\end{minipage}
}
\hspace{1.2em}
\subfloat[
\textbf{Resolution Matters.} 
\ourmethod~does worse than region encoder-only given no context, but performs much better with multiple regions while using the same \# of parameters.
\label{tab:resolution}
]{
\centering
\begin{minipage}{0.30\linewidth}{\begin{center}
\tablestyle{4pt}{1.05}
\begin{tabular}{cccc}
model & size & acc & params \\
\specialrule{.15em}{.075em}{.075em} 
\multirow{3}{*}{Swin-T} & 256 & 53.76 & 31M \\ 
                        & 256/256 & 52.93 & 47M \\ 
                        & \cellcolor{lightblue}512/256 & \textbf{60.56} & 47M \\ \hline
\multirow{3}{*}{Swin-L} & 256 & 68.78 & 206M \\ 
                        & 256/256 & 67.84 & 215M \\ 
                        & \cellcolor{lightblue}512/256 & \textbf{72.42} & 215M 
    \vspace{1em}
\end{tabular}%
\end{center}}
\end{minipage}
}
\label{tab:ablations} 
\caption{\textbf{Ablating \ourmethod~design choices.}
We highlight our defaults in \colorbox{lightblue}{blue}~and \textbf{bold} the best numbers.
}
\end{table*}

\subsection{Object Detection}

We use SwinV2 \cite{liu2021swinv2} as the backbone and adopt DetINO \cite{zhang2022dino} for the detection head. We follow the standard COCO \cite{linMicrosoftCOCOCommon2015} 1x training schedule and train on Cityscapes for 12 epochs with the AdamW optimizer  ($\beta_1=0.9$, $\beta_2=0.999$) and a learning rate of $1\times10^{-3}$. The learning rate for the SwinV2 backbone is scaled by a factor of 0.1. The input is first padded to the nearest multiple of chip size first, and then chipped in the same way as \cref{sec:cls}. Each region is then processed by the region encoder independently. The outputs from Swin encoder are concatenated to form a global feature map of the entire image, which is then processed by the context encoder. The resulting sequence is passed to the DetINO detection head.

We report $\text{mAP}$ for our detection experiments. These experiments are relatively brief---na\"ive usage of the Swin backbone results in out-of-memory errors for the large 1,024 x 2,048 images found in Cityscapes. Conversely, \ourmethod~is able to model the images with ease.

\section{Ablations}
\label{sec:ablations}
In this section, we ablate our design choices based on iNaturalist classification, particularly around our context encoder. 
The settings that we keep constant are that \ourmethod~is run on 512/256 inputs, where we chunk our four regions together to create one sequence into our context encoder. 
As this usually fits in memory, we have no need to perform cross-chunk contextualizing as we do in xView. 

\subsection{Context Encoder Architecture}
We tested ViT, Hyper, and Mamba as context encoders in the course of our classification experiments. These results are detailed \Cref{tab:ablation_context_backbone}. iNaturalist 2018 resizes their images to $800\times 800$ pixels for training. These images can be modeled comfortably by ViT and benefit from full self-attention. However, both Hyper and Mamba both perform better than ViT as context encoders with the added benefit of having a much larger capacity for scale.

While Mamba has less parameters than both ViT and Hyper---up to 8\% fewer parameters for Swin-T \xT Mamba than for Swin-T \xT Hyper---this difference disappears as the region encoder increases in size as seen in \Cref{tab:results_inat}. Furthermore, the decrease in parameters ultimately has an insignificant impact on the peak memory usage of the overall \ourmethod~model.

\subsection{Context Encoder Depth}
Crucial to our design is how deep our context encoder should be, as our goal is to keep it as lightweight as possible so its overhead is minimal. 
We find that an acceptable increase in parameters keeps the number of layers to either 1 or 2. 
As shown in Table~\ref{tab:context_encoder_depth}, larger region encoders generally benefit from having deeper context encoders.
The accuracy is the greatest when the depth is 2 for the largest model, and the trade-off is acceptable for the smallest model, being within 1 accuracy point, so we choose depth 2 as our default. 

\subsection{Resolution Size Matters}
We revisit the central assumption of our work, which is if high resolution is essential for understanding. 
In Table~\ref{tab:resolution}, we emphasize our results on iNaturalist which prove precisely that models previously did not take advantage of such context, and now do with \ourmethod.
Comparing the Swin-T/L 256 run with Swin-T/L 256/256, which is our method taking in no context, our model actually does worse with extra parameters, likely due to non-functional parameters interfering with the model's learning.
However, once we increase our input image size by $4\times$ to $512/256$, our model is immediately able to take advantage of the context with no increase in parameters, boosting accuracy up by 4.6\%-7.6\%. 

\section{Discussion}
\label{sec:discussion}

\subsection{Effective Receptive Field}
\label{sec:ablation-erf}

\begin{figure}
    \centering
    \includegraphics[width=\columnwidth]{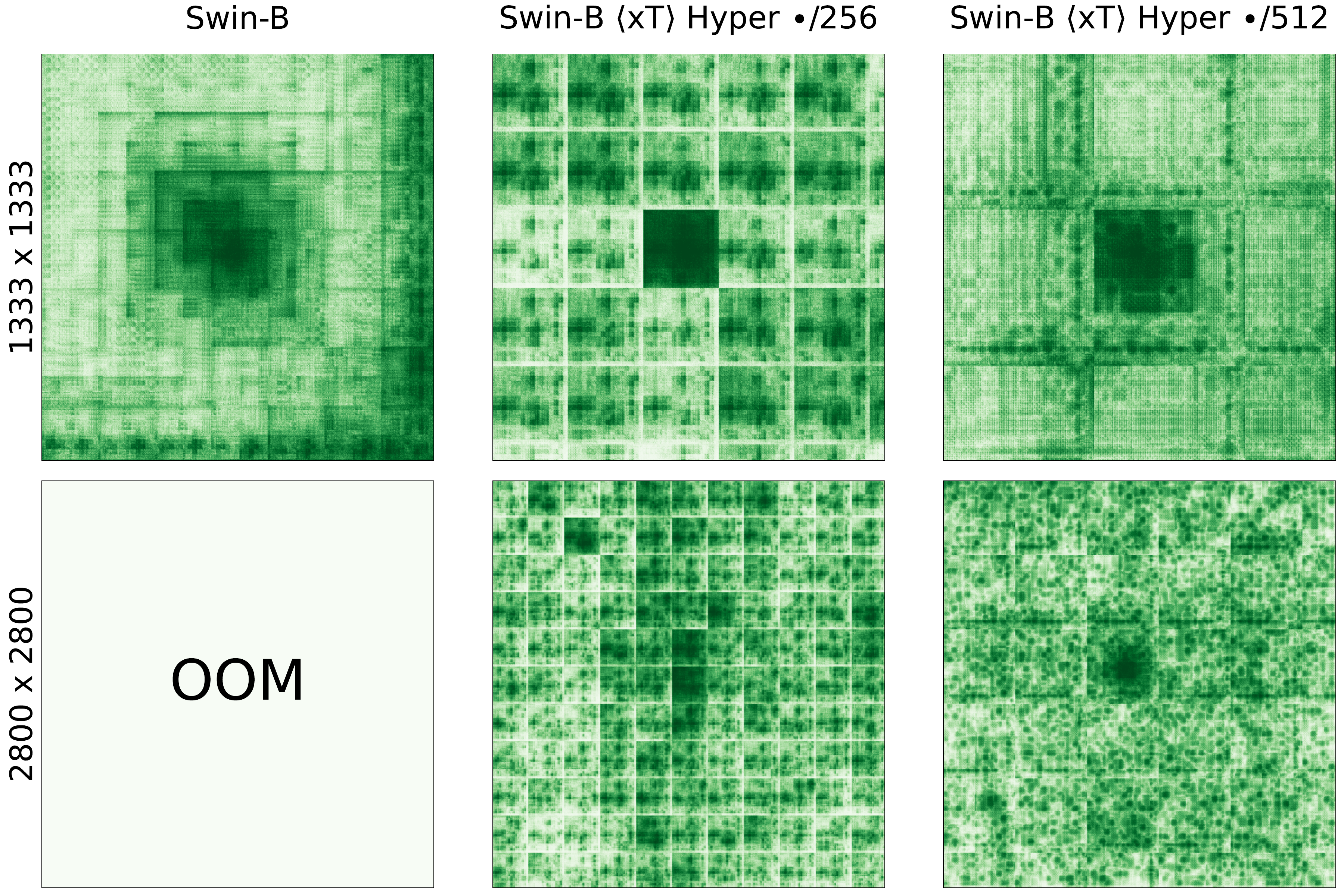}
    \caption{\textbf{Effective receptive fields of Swin-B and Swin-B \xT Hyper}. The center feature from the last layer of region encoder is used to assess sensitivity to areas across the image. Darker green signifies greater sensitivity. 
    }
    \vspace{-1em}
    \label{fig:erf}
\end{figure}

We empirically visualize the effective receptive field (ERF) \cite{luo2016understanding} of \ourmethod~ in \cref{fig:erf}.
The effective receptive field is computed by setting the gradient of one pixel in the final feature map to 1 while keeping the gradient of all other pixels as zero, and then computing the gradient with respect to the input image through a backward pass. 
The visualized image measures the magnitude of gradient with darker green areas demonstrating greater ``sensitivity'' and lighter areas demonstrating lower. 
We show that Swin Transformer has an ERF similar to a skewed Gaussian which vanishes quickly over distance. Comparatively, \ourmethod~retains a more uniform distribution across the entire image. Since our nested tokenization approach is effectively a convolution, we see convolution-like artifacts in the ERF that are mitigated as region sizes get larger.
This highlights the \ourmethod~framework's capacity at capturing and integrating long range context in large images.

\subsection{Throughput}
\begin{figure}[!t]
    \centering
    \includegraphics[width=\columnwidth]{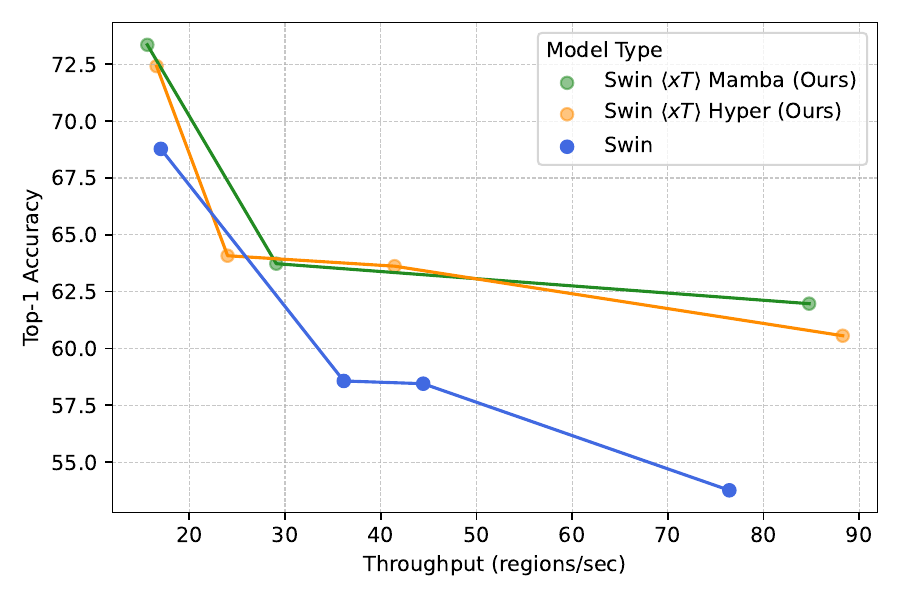}
    \vspace{-2em}
    \caption{\textbf{\ourmethod~offers greatly increased accuracy per throughput.}  
    On iNaturalist classification, we find that our models only slightly diminished throughput (with the exception of Swin-T \xT XL/Mamba) but achieved greater accuracies at each throughput threshold.}
    \label{fig:throughput}
\end{figure}

In lieu of FLOPS, which is difficult to standardize, hardware dependent, and an inconsistent measure of running time when using custom kernels, we instead report the throughput of our method and compare to prior works in Figure~\ref{fig:throughput}. 
For our comparisons, we use 40GB Nvidia A100 GPUs.
As we work with multiple image resolutions at once, we calculate a throughput of \emph{regions/second}, which is the number of encoder-sized regions we process per second. 

On a per-model size comparison, our method drops in throughput slightly compared to prior methods (with the exception of the fastest size, Tiny). 
However, we are able to achieve large accuracy gains for this slight tradeoff which are much better than if we used a larger model from prior work instead. 
This demonstrates the benefit of our method in improving the frontier in modeling larger images efficiently.

\begin{figure}[ht]
    \centering
    \includegraphics[width=\columnwidth]{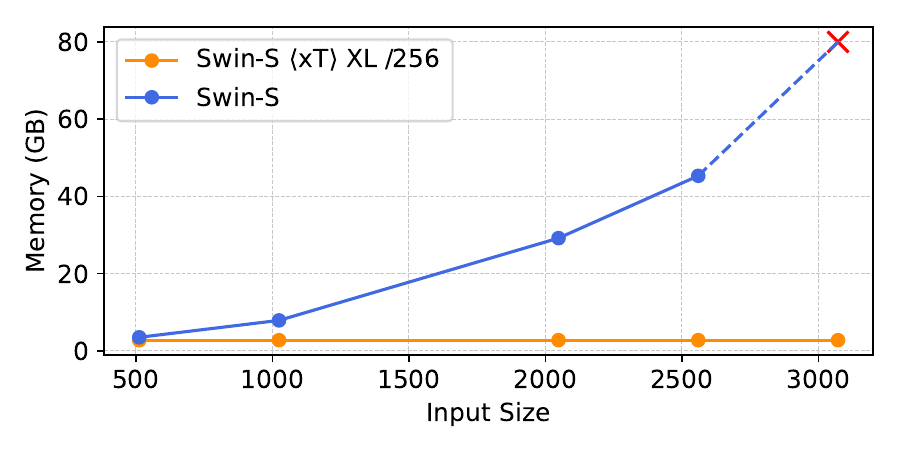}
    \vspace{-2em}
    \caption{\textbf{Swin rapidly goes out of memory (indicated by the red {\color{red}{X}}) as images grow in size whereas \ourmethod~retains near-constant memory cost}. \ourmethod~can scale to much larger images than the na\"ive usage of a vision backbone.}
    \label{fig:mem}
\end{figure}

\subsection{Memory Growth}
Modern vision backbone, such as Swin, rapidly run out-of-memory near-quadratically as the image size increases. In \Cref{fig:mem}, we show how the \ourmethod~framework effectively removes this quadratic memory growth, allowing the model to reason over large images end-to-end.
Thanks to our chunking design, our method incurs minimal extra memory when we increase our input size as we recurrently store previous regions as context for future context-aware encoding. 

\subsection{Locality of Features}
Many vision backbones have adjustable receptive fields for the integration of local information. Larger receptive fields have memory and throughput implications, however. We ablate the Swin transformer for various region and window sizes and compare the impact on accuracy, throughput, and memory in \Cref{tab:ablation_window_size} on the iNaturalist classification task.

We set the window size used by Swin appropriately as a function of region size. That is, an region size of 256 corresponds to a window size of 16, region size of 512 to a window size of 32, and so on. Swin-B, without the usage of \ourmethod, can achieve marginally higher accuracy at small image sizes. However, this trend is reversed when working with larger images. Of importance is the ability for \ourmethod~to retain high throughput (measured as time per epoch in minutes) and low memory usage (memory per region in GB) while achieving equivalent accuracy downstream.

\setlength{\arrayrulewidth}{0.8pt}
\begin{table}
    \centering
    \resizebox{\columnwidth}{!}{%
    \begin{tabular}{l|c|r|r|rr}
    Model & Acc. & Size(s) & W. Size & Th. & Mem.\\ 
    \specialrule{.15em}{.075em}{.075em}
    Swin-B & \textbf{67.02} 	& 512 & 32 	& 29.53 & 6.02 \\
    \rowcolor{lightblue}
    Swin-B \xT XL &	65.49 & 512/256 & 16 	& \textbf{12.77} & \textbf{0.49} \\
    \hline
    Swin-B & 67.37 	& 1024 	& 64 & 242.92 & 25.1 \\
    \rowcolor{lightblue}
    Swin-B \xT XL &	\textbf{68.19} 	& 1024/256 	& 16 & \textbf{64.7} & \textbf{2.03} \\

    \end{tabular}
    }
    \caption{\textbf{Comparison of top-1 accuracy (Acc.), time per epoch (Th.), and memory per region (Mem.) as a function of region size(s) and window sizes.} 
    Swin-B and \colorbox{lightblue}{our method} have approximately equivalent accuracies, but \ourmethod~achieves much more desirable throughput and memory utilization.}
    \label{tab:ablation_window_size}
\end{table}

\section{Conclusion}
\label{sec:conclusion}

Large images have been the norm for the past decade in consumer and professional domains. Yet, state-of-the-art methods in computer vision still limit themselves to modeling small images by throwing away valuable information via down-sampling and cropping. Large images from the real world contain a wealth of information that often require context across long distances to properly make sense of.

In this work, we introduce \ourmethod, a framework through which vision models trained to handle small images can cheaply integrate larger context over large images. \ourmethod~achieves significant gains in downstream tasks such as classification, detection, and segmentation on real-world datasets. By offering a simple framework by which vision researchers can effectively model large images, we aim to widen the types of data and tasks vision researchers aim to solve.

\section*{Impact Statement}
This paper presents work whose goal is to advance the field of machine learning. There are many potential societal consequences of our work, none which we feel must be specifically highlighted here.

\section*{Acknowledgements}
Trevor Darrell’s group was supported in part by funding from the Department of Defense as well as BAIR’s industrial alliance programs. This work was additionally supported by the DoD's High Performance Computing Modernization Program. Ritwik Gupta is supported by the National Science Foundation under Grant No. DGE-2125913 and funding from the UC Berkeley AI Policy Hub.
We are grateful to Bryan Comstock for his timely technical support even when the authors were asking a lot of questions. Thank you to Suzie Petryk, Mark Lowell, and William Yang for reviewing the transcript.

\clearpage
\bibliography{main,refs}
\bibliographystyle{ieeenat_fullname}



\end{document}